\documentclass[11pt,a4paper]{article}
\usepackage[hyperref]{acl2020}
\usepackage{times}
\usepackage{latexsym}
\usepackage{enumerate}
\usepackage{hyperref}

\usepackage{microtype}
\usepackage{graphicx}

\aclfinalcopy % Uncomment this line for the final submission
\title{Aspect-based Sentiment Analysis in Document - FOMC Meeting Minutes on Economic Projection}

\author{Yifei Wang \\
  University of California,Berkeley  \\
  \texttt{sarahwang688@berkeley.edu} \\}

\date{}

\begin{document}
\maketitle
\begin{abstract}
The Federal Open Market Committee (FOMC)[1] within the Federal Reserve System is responsible for managing inflation, maximizing employment, and stabilizing interest rates. Meeting minutes play an important role for market movements because they provide the bird’s eye view of how this economic complexity is constantly re-weighed. Therefore, There has been growing interest in analyzing and extracting sentiments on various aspects from large financial texts for economic projection. However, Aspect-based Sentiment Analysis (ABSA) is not widely used on financial data due to the lack of large labeled dataset. In this paper, I propose a model to train ABSA on financial documents under weak supervision and analyze its predictive power on various macroeconomic indicators.
\end{abstract}

\section{Introduction}

There has been growing interest in analyzing and extracting sentiments from large financial texts for economic projection.  The Federal Open Market Committee (FOMC)[1] within the Federal Reserve System is responsible for managing inflation, maximizing employment, and stabilizing interest rates. Committee members meet eight times every year and meeting minutes are usually released to the public shortly after the meeting. Meeting minutes play an important role for market movements because they provide the bird’s eye view of how this economic complexity is constantly re-weighed. How hawkish or dovish are the communications? How do these opinions change over time? What measure of consensus is there? The tone, tolerance and diversity of the minutes can have an impact on markets and trading. However, ABSA is not widely used on financial data due to the lack of a large labeled dataset. In this paper, I want to answer the following questions - Without much labeled data, are we able to dynamically extract aspects in a weakly supervised way, and can aspect-based sentiment extracted from FOMC meeting minutes be used for economic projection. To answer the questions, I will extend sentence-level natural language processing models to document-level.

Aspect-based sentiment analysis on financial data is less explored due to the lack of labeled financial sentiment data set. And, as mentioned above, committee members usually discuss various topics (aspects) related to the economy, such as inflation, growth, employment, etc. Therefore, the goal of the model is to be able to extract aspects without much labeled data. The second challenge lies in the text and tone nuances in FOMC meeting minutes. It is uncertain whether the models is able to understand the pragmatics of documents in order to pick up committee members’ intention. Last but not the least, another challenge is the large document data processing across the full sample’s time series. Currently NLP models mostly focus on word-level or sentence-level analysis. Therefore, we need to explore how expensive it is to expand to lengthy document-level analysis.

\section{Related Work}
There has been some work related to sentence analysis that can be expanded to document level. For example, in [2], BERT advances the state-of-art results for sentence level NLP tasks by jointly analyzing both left and right text in all layers. Author in [3] further develops the FinBERT model by training BERT on large financial corpora, which makes FinBERT able to outperform the state-of-art results in many domain-specific tasks related to finance.

There has also been some literature related to Aspect-Based Sentiment Analysis (ABSA).  Paper in [4] shows the potential of using the contextual word representations from the pre-trained language model BERT, together with a fine-tuning method with additional generated text, in order to solve out-of-domain ABSA and outperform previous state-of-the-art results on several NLP tasks.

On the financial side, the author in [5]  applied pre-trained BERT on aspect classification and sentiment prediction on labeled financial data provided by Financial Opinion Mining and Question Answering Open Challenge held at WWW 2018 Lyon, France. [6] applies Latent Semantic Analysis to show that themes extracted from FOMC meeting minutes are correlated with current and future economic conditions. In [7], the author uses textual information from FOMC meeting minutes to predict directional movement for fed fund rate.

However, to the best of my knowledge, there has not been any research on extracting aspects from FOMC meeting minutes without much labeled dataset and applying sentiment prediction for economic projection.

\section{Methods}
\subsection{Document Data}
FOMC meeting minutes have been available publicly online since 1939. I used the last 10 years as my sampling period. Data is pre-processed by separating each document into a list of sentences using the NLTK package. I further converted each sentence to lower cases, and removed sentences only containing numbers but no texts. Sentences that repeatedly appear in the document but no financial information, such as “return to the previous page”, are removed as well. Sentences with less than 7 words are removed and sentences with more than 80 words are truncated. Here are some summary statistics for the pre-processed data:
\begin{itemize}
\item Number of documents: 480
\item Number of words: 160,098,864
\item Maximum / median / minimum length of document (in words): 10663 / 3447 / 303
\item Maximum/ median / minimum length of sentence (in words): 80 / 38 / 7
\end{itemize}

\subsection{Macroeconomics Data}
I will use Inflation, GDP growth, and unemployment rate retrieved from Bloomberg, to assess FOMC meeting minutes’ predictive power on macroeconomic data. Bloomberg data is updated at daily frequency and they are aggregated to monthly frequency for analysis.

\subsection{Models}
\subsubsection{Embeddings}
FinBERT is a pre-trained NLP model to analyze sentiment of financial text. It is built by further training the BERT language model in the finance domain, using the Financial PhraseBank created by Malo et al. (2014) [8] and FiQA Task 1 sentiment scoring dataset [9]. Embeddings are crucial in analyzing ABSA. Therefore, I explored 3 models to extract embeddings:
\begin{itemize}
\item The base model is to use word embeddings from FinBERT by averaging the last 4 layers from FinBERT.
\item The second model is to extract sentence embeddings by averaging pooling word tokens. By comparing the first and second model, I hope to get a better understanding whether word embeddings or sentence embeddings fit better for my research purpose considering both accuracy and efficiency.
\item After narrowing down to whether to use word or sentence embeddings from model 1 and 2, the third model is to explore whether we can achieve better prediction results by further fine-tuning the FinBERT model.
\end{itemize}
\subsubsection{Fine-Tuning FinBERT Model}
FinBERT is trained on Financial PhraseBank. Financial PhraseBank consists of 4845 sentences randomly selected from financial news from LexisNexis database. Each sentence is labeled by 16 annotators with background in finance and business, based on their judgement whether the given sentence will impact stock price in a positive, negative or neutral direction. Table below is the distribution of sentiment labels and agreement level for Financial PhraseBank’s sentences. FinBERT model is trained on all 4845 sentences. For fine-tuning purposes, I only used those sentences with 100\% agreement level, which takes up around 47\% of total sample size, to reduce noises introduced by data labelling.  I lowered the learning rate to be 1e-6 and added a dropout rate to be 0.1. The model is trained on 10 epochs. 
\begin{figure}[h]
\includegraphics[width=0.5\textwidth]{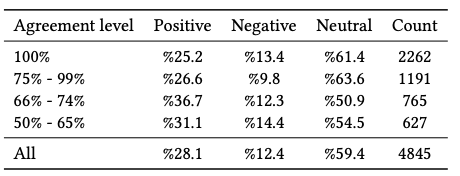}
\caption{Distribution of Sentiment Labels and Agreement Level}
\end{figure}

\subsubsection{Aspect Classification}
I predefined aspects as inflation, growth, and employment, because they are widely monitored economic themes in the market. With embeddings from above, the model will calculate a one-hot vector for each of the predefined aspects, and sentence vector for each sentence in the documents. Distances between aspect and each sentence are calculated using cosine similarity. Final aspect is the one with highest cosine similarity.

\subsubsection{Sentiment Prediction}
After aspect classification, we will analyze and predict each aspect’s sentiment separately. The reason for doing this is that there may be mixed views for aspects in one document. For example, one speech could be hawkish on inflation but not necessary on employment due to technological structural change. Secondly, for each of the predefined aspects, we use a dedicated economic indicator to analyze the predictive power from sentiment analysis. Sentiment prediction is modeled on both FinBERT model and fine-tuned FinBERT model. Sentiment classification is conducted by adding a dense layer after the last hidden state of the [CLS] token. This is the recommended practice for using BERT for any classification task [10]. 

\section{Results}
From initial analysis using FinBERT’s sentence and word embeddings, I found that the distribution of aspect classification is more well balanced using sentence embeddings than using word embedding, as shown in Figure 2. The reason might be because word embeddings from FinBERT are context related. It is not the best way to analyze each individual word separately without considering the sentence’s context. Therefore, I narrowed down to use sentence embeddings for the rest of the experiment.
\begin{figure}[h]
\includegraphics[width=0.5\textwidth]{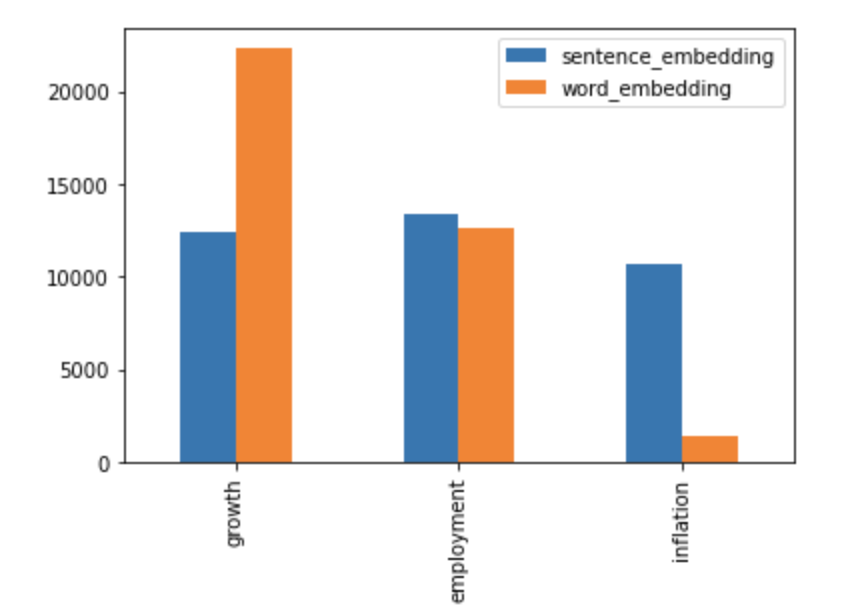}
\caption{Number of Observations for Each Aspect}
\end{figure}
During fine-tuning, the sample is split into 80/20 between training and test datasets. After fine-tuning FinBERT model, we are able to achieve the accuracy level of 0.98, which is improved from 0.97 from the original paper [3], and loss is improved from 0.13 in [3] to 0.06. Results are shown in Figure 3.

\begin{figure}[h]
\includegraphics[width=0.5\textwidth]{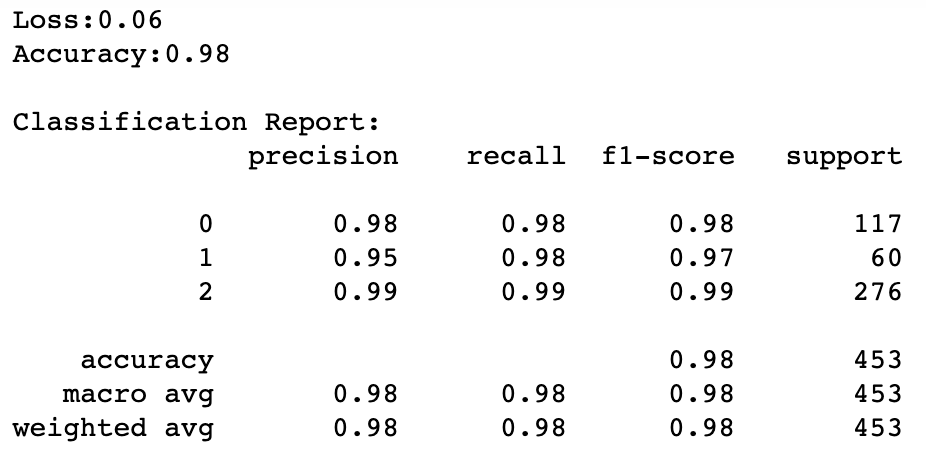}
\caption{Classification Report for Fine-Tuned FinBERT Model}
\end{figure}
The relationship between FOMC meeting minutes and macroeconomic data are assessed through statistical significance analysis from Figure 4 to Figure 6. Our growth sentiment is able to explain around 63\% growth rate change in the market, and employment sentiment and inflation is able to explain around 47\% and 19\% of their respective macroeconomic variables. All three OLS regressions show the results are statistically significant.
\newpage

\begin{figure}[h]
\includegraphics[width=0.5\textwidth]{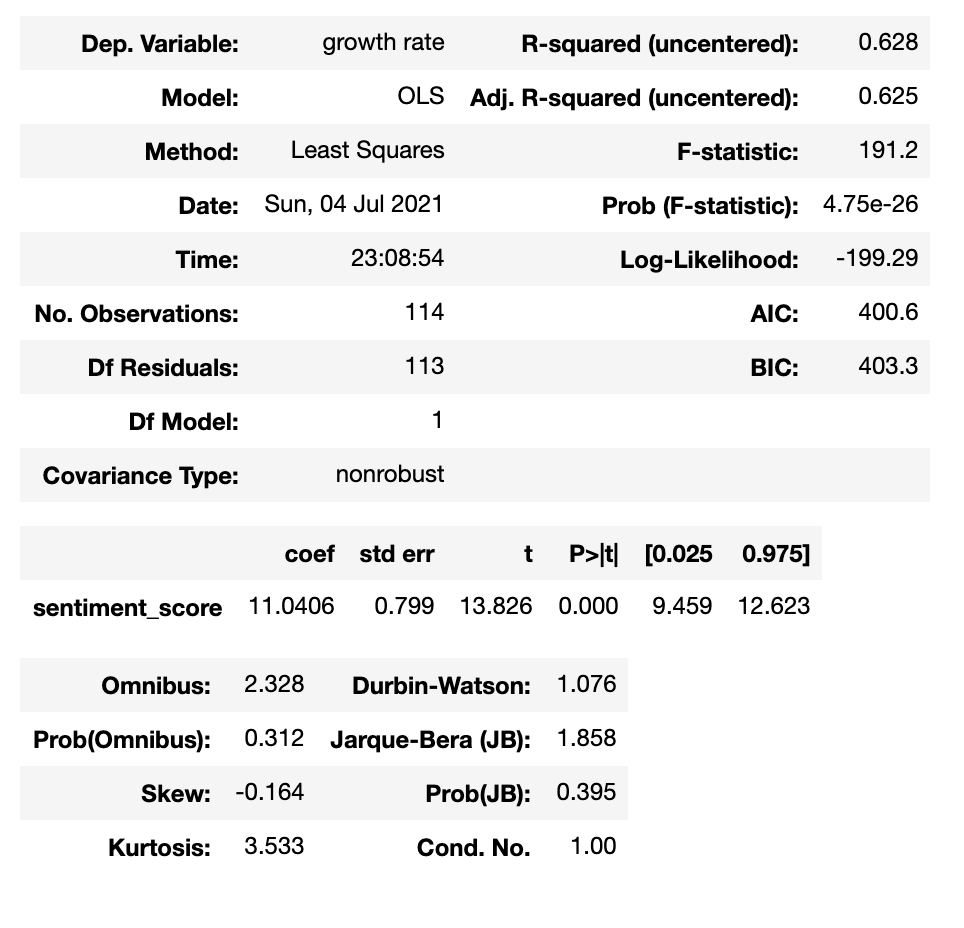}
\caption{OLS Regression Result on Growth Rate and Growth Sentiment}
\end{figure}

\begin{figure}[h]
\includegraphics[width=0.5\textwidth]{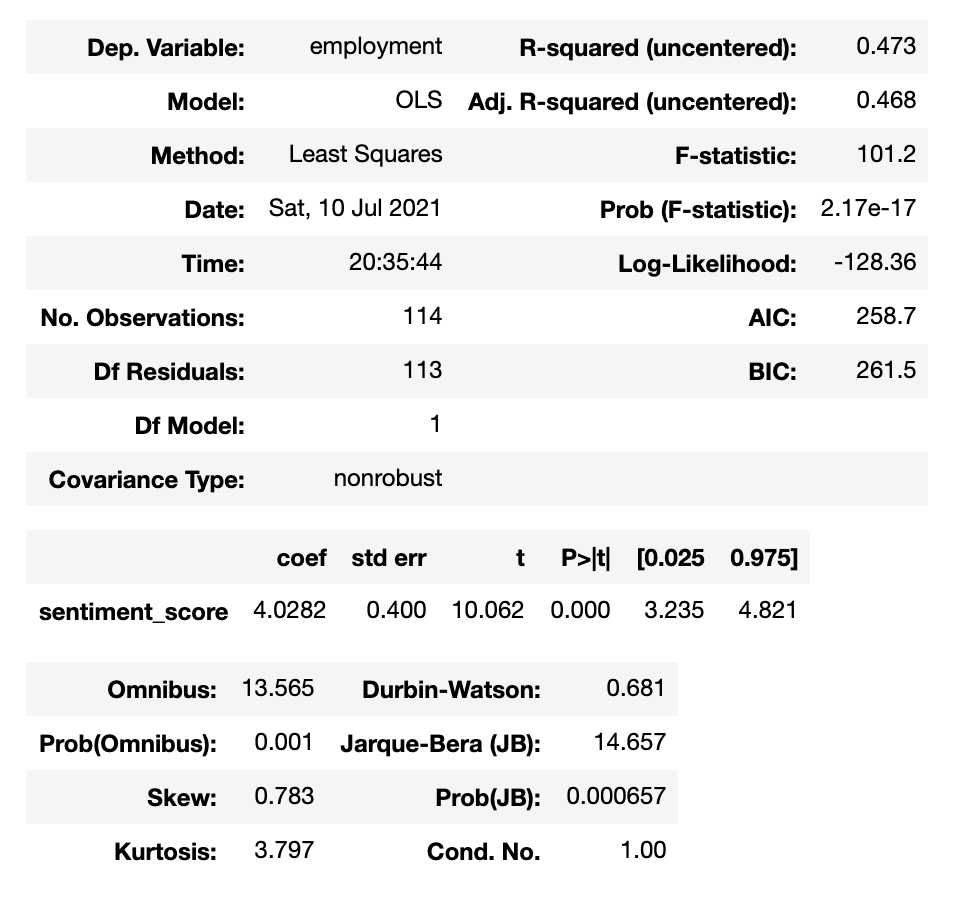}
\caption{OLS Regression Result on Employment and Employment Sentiment}
\end{figure}

\begin{figure}[h]
\includegraphics[width=0.5\textwidth]{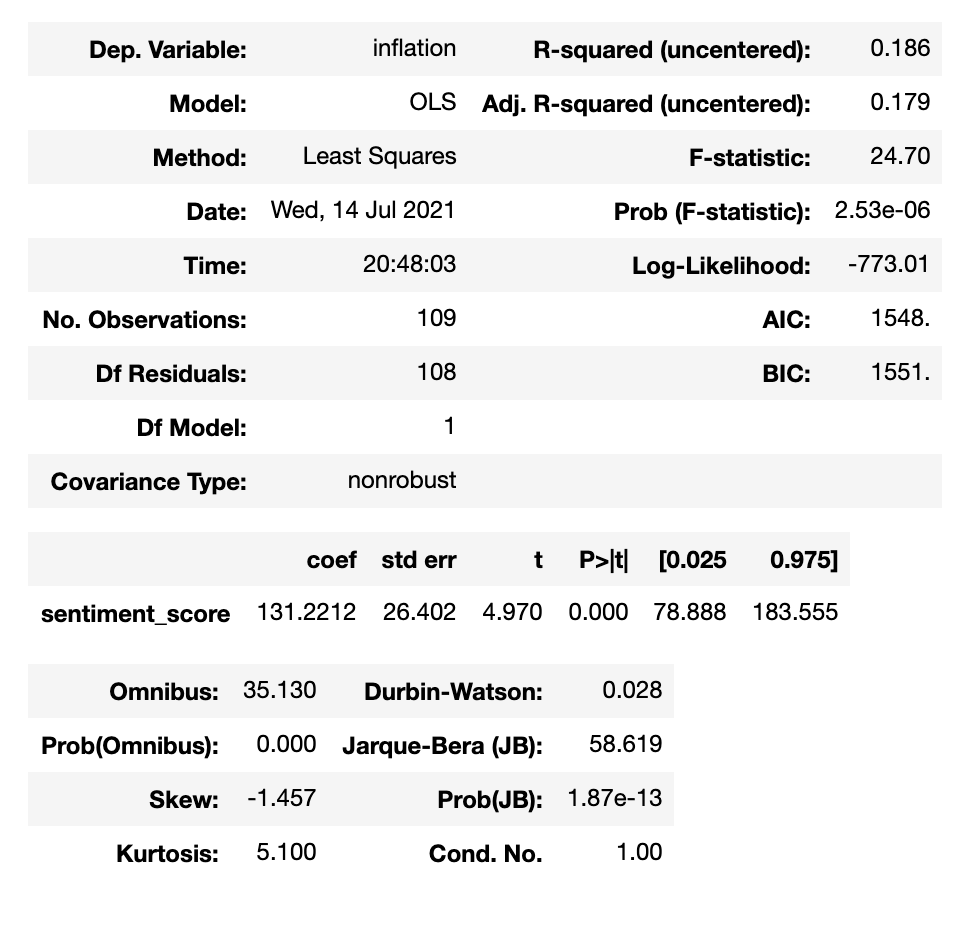}
\caption{OLS Regression Result on Inflation and Inflation Sentiment}
\end{figure}
\newpage
\section{Conclusion}
In this paper, I proposed a model to extract aspects under weak supervision to address one of the limitations in ABSA application in financial analysis due to the lack of labeled dataset. Aspect-based sentiment predictions are further regressed on market macroeconomic data. It is found that textual information does have predictive power on economic performance and OLS regression results are statistically significant. 

\newpage

\section{Reference}

\begin{enumerate}
\item Federal Open Market Committee. Background on Policy Statements.\\ \url{https://www.federalreserve.gov/monetarypolicy/fomc_historical.htm}
\item Jacob Devlin, Ming-Wei Chang, Kenton Lee, and  Kristina Toutanova. BERT: Pre-training of Deep Bidirectional Transformers for Language Understanding. Proceedings of the 2019 Conference of the North American Chapter of the Association for Computational Linguistics: Human Language Technologies, Volume 1
\item Dogu Tan Araci. FinBERT: Financial Sentiment Analysis with Pre-trained
Language Models.\\
\url {https://arxiv.org/pdf/1908.10063.pdf}
\item Mickel Hoang, Oskar Alija Bihorac. Aspect-Based Sentiment Analysis Using BERT.
\url {https://aclanthology.org/W19-6120.pdf}
\item Ashish Salunkhe, Shubham Mhaske. Aspect Based Sentiment Analysis on Financial Data using Transferred Learning Approach using Pre-Trained BERT and Regressor Model
\url {https://www.irjet.net/archives/V6/i12/IRJET-V6I12179.pdf}
\item Ellyn Boukus and Joshua V. Rosenberg. The Information Content of FOMC Minutes. SSRN Electronic Journal, 2011.
\item Ye Ye. Can FOMC Minutes Predict the Federal Funds Rate? Stanford University
\item Pekka Malo, Ankur Sinha, Pekka Korhonen, Jyrki Wallenius, and Pyry Takala.
2014. Good debt or bad debt: Detecting semantic orientations in economic texts.
Journal of the Association for Information Science and Technology 65, 4 (2014),
782–796. 
\url {https://doi.org/10.1002/asi.23062 arXiv:arXiv:1307.5336v2}
\item Macedo Maia, Siegfried Handschuh, André Freitas, Brian Davis, Ross Mcdermott,
Manel Zarrouk, Alexandra Balahur, and Ross Mc-Dermott. 2018. Companion of
the The Web Conference 2018 on The Web Conference 2018, {WWW} 2018, Lyon
, France, April 23-27, 2018. ACM. 
\url {https://doi.org/10.1145/3184558}
\item Jacob Devlin, Ming-Wei Chang, Kenton Lee, and Kristina Toutanova. 2018. BERT: Pre-training of Deep Bidirectional Transformers for Language Understanding. (2018). 
\url{https://doi.org/arXiv:1811.03600v2 arXiv:1810.04805}
\end{enumerate}

\end{document}